\documentclass[conference,onecolumn]{IEEEtran}
\IEEEoverridecommandlockouts

\usepackage{cite}
\usepackage{amsmath,amssymb,amsfonts}
\usepackage{graphicx}
\usepackage{booktabs}
\usepackage{microtype}
\usepackage{xcolor}
\usepackage{tikz}
\usetikzlibrary{arrows.meta,positioning,shapes.geometric,fit}
\usepackage[colorlinks=true,allcolors=blue]{hyperref}

\newcommand{\src}{\texttt{source.\$in}}

\begin{document}

\title{Probe, Don't Prompt: A Hidden-State Probe\\
for Metadata Filtering in Multi-Meta-RAG}

\author{\IEEEauthorblockN{Mykhailo Poliakov and Nadiya Shvai}
\IEEEauthorblockA{\textit{National University of Kyiv-Mohyla Academy}\\
Kyiv, Ukraine\\
\texttt{mykhailo.poliakov@ukma.edu.ua}, \texttt{n.shvay@ukma.edu.ua}}}

\maketitle

\begin{abstract}
Multi-Meta-RAG improves retrieval for multi-hop question answering by filtering
a vector store on metadata, the news source and the publication date, which it
extracts from each query with GPT-3.5. This extractor requires a
proprietary API call for every query, and its output is not restricted to the allowed
sources. Even though the prompt specifies 49 sources, GPT-3.5 generates 128 distinct
source names, 79 of which do not appear on the list. We show that the source part of
this extractor can be replaced by a local, deterministic probe trained on the
hidden states of a small open-source language model. The probe is a
multi-label logistic classifier on the mean-pooled hidden states of one layer,
trained with a class-balanced loss on the gold evidence sources. Using 5-fold
cross-validation to select the layer, we evaluate four models ranging from 135M
to 1.5B parameters out-of-fold across all 2556 MultiHop-RAG queries. The probe reaches a
set-exact accuracy of 90.9\%, compared to 88.0\% for a substring baseline and
80.9\% for GPT-3.5. The difference comes from null queries, for which GPT-3.5
never predicts an empty source set. Across the remaining queries, all three methods achieve set-exact accuracies
within 1.1 percentage points of each other. Unlike earlier probing work, we use
the probe as a component of a retrieval pipeline rather than for interpretation.
For all four models, the cross-validation selects one of the first layers, in contrast to the middle layers reported for retrieval
tasks. We attribute this to the source name appearing verbatim in 95.4\%
of gold query--source pairs. The 135M model is 1.5 percentage points less accurate than
the 1.5B model, so the filter requires only a partial
forward pass through a small model and one linear layer. The code is available
at \url{https://github.com/mxpoliakov/Multi-Meta-RAG}.
\end{abstract}


\section{Introduction}
Retrieval-augmented generation (RAG)~\cite{lewis:2020} grounds a language
model on documents retrieved from an external store. However, RAG performs
poorly on \emph{multi-hop} queries, which require evidence from several
documents at once. The MultiHop-RAG benchmark~\cite{tang:2024} shows this problem on 2556 news queries whose answers combine two to four articles.
Multi-Meta-RAG~\cite{poliakov:2024} improves multi-hop retrieval on this
benchmark by filtering the vector store on metadata before the similarity
search. The main filter field is the news source named in the query (e.g.\
\emph{The Verge} or \emph{TechCrunch}), which is extracted by prompting
GPT-3.5\footnote{\texttt{gpt-3.5-turbo-1106}}.

This extractor has two drawbacks. Firstly, it requires a call to a proprietary
API for every query, which adds token cost and network latency to retrieval.
Secondly, its output is not restricted to the allowed sources. Although the prompt lists 49 sources, GPT-3.5 produces 128 distinct strings across the dataset, 79 of which do not belong to the allowed set. Many of its filters therefore refer to
sources that the vector store does not contain.

Our starting point is that the source is almost always named in the query
itself. In 95.4\% of gold query--source pairs, the source name appears verbatim within the query text. Extracting the source therefore does not require
generation, and we expect the information to be linearly available in the
hidden states of a language model. We train a lightweight
\emph{probe}~\cite{alain:2016} on the hidden states of a small open-source model,
following evidence that intermediate representations carry task-relevant signal
for retrieval~\cite{lin:2025}. The probe is a multi-label classifier over a fixed vocabulary. Its output space consists of the 49 allowed sources, so, unlike the
prompted extractor, it cannot produce a source outside the list.

Our primary contributions are as follows.
\begin{enumerate}
\item We design a fixed-vocabulary probe that replaces GPT-3.5 source extraction in Multi-Meta-RAG while leaving date filters unchanged. It requires no
API call, cannot produce a source outside the allow-list, and is deterministic.
\item We compare the probe with GPT-3.5 and with a model-free substring baseline
on all 2556 MultiHop-RAG queries using set-exact accuracy. The probe reaches 90.9\%, compared to 88.0\% for the substring baseline and 80.9\% for GPT-3.5. The
difference comes from null queries, for which the probe predicts an empty set. On
the other queries, it is within 1.1 percentage points of both baselines.
\item We show that cross-validation consistently selects an early layer rather than an intermediate or final layer, and that mean pooling outperforms last-token pooling across all four models.
\item We compare four models from 135M to 1.5B parameters. The 135M model is 1.5
percentage points less accurate than the 1.5B model, so the filter can be
computed with a very small model.
\end{enumerate}

The remainder of this article is organized as follows. We first review related work.
Then, we describe the probe and the choices that its accuracy depends on. Next, we benchmark the probe against GPT-3.5 and the substring baseline, examine the effects of layer depth, pooling, and model size, and relate our findings to prior work.
Finally, we draw conclusions and discuss limitations and future work. The code
is available at \url{https://github.com/mxpoliakov/Multi-Meta-RAG}.

\section{Related Work}
\subsection{Metadata Filtering and Entity Extraction}
While standard RAG~\cite{lewis:2020} using dense retrieval~\cite{karpukhin:2020} struggles on multi-hop queries~\cite{tang:2024}, metadata filtering with LLM-extracted constraints improves retrieval~\cite{poliakov:2024}. Extracting entities from the query and
filtering on chunk metadata is a routing step of advanced RAG
pipelines~\cite{gao:2023}, and LLMs are also used to rewrite the query itself
before retrieval~\cite{ma:2023}. In these pipelines, the extractor is typically
a prompted LLM that generates free-form text.

Several approaches avoid free-form generation. Constrained decoding restricts
a generator to a fixed output set, with a grammar~\cite{geng:2023} or a prefix
tree over the allowed names~\cite{decao:2021}. Span-based named-entity
recognisers such as GLiNER~\cite{zaratiana:2024} extract entities without an
LLM. Another approach distills the prompted model~\cite{hinton:2015} by training a smaller network on generated outputs, as demonstrated by UniversalNER using ChatGPT for open named-entity recognition~\cite{zhou:2024}.
We instead train a small local model directly on the gold evidence sources.
Capable small open-source models~\cite{qwen:2024,benallal:2025} and the finding
that decoder hidden states are strong text features~\cite{behnamghader:2024} make
this practical.

\subsection{Probing Classifiers and Layer Selection}
Probing classifiers trained on frozen hidden states show what information these representations encode~\cite{alain:2016,conneau:2018}, as reviewed in~\cite{belinkov:2022}. Rigorous probing separates what a
probe learns from what can be recovered from the surface form
alone~\cite{hewitt:2019}. In our study, the model-free substring baseline plays
this control role. Hidden-state representations have likewise been probed for monitoring model behavior, such as classifying whether a given statement is factual~\cite{azaria:2023}. We
use a probe neither for interpretation nor for monitoring, but as a component of
a retrieval pipeline.

Layer-wise probing shows that different kinds of information are found at
different depths of the network~\cite{tenney:2019}. Intermediate layers often outperform the last layer in downstream tasks~\cite{skean:2025}, including multi-hop retrieval~\cite{lin:2025}. We test this for source extraction. In our
experiments, the first layers outperform the middle ones, which we attribute to
the source name being present in the surface form of the query rather than in
its deeper semantics.

\section{Method}
This section describes the probe. We first define the task and the evaluation
metric. We then describe the pipeline and the choices that the accuracy of the
probe depends on. Finally, we outline the baseline methods.

\subsection{Task and Data}
We use all 2556 MultiHop-RAG queries, of which 856 are comparison, 816
inference, 583 temporal and 301 null queries. The label space consists of the 49
sources of the Multi-Meta-RAG prompt allow-list. The formulation is multi-label, since queries reference between one and four sources (598 queries refer to one source, 1340 to two, 300 to three, and 17 to four). Let $\mathcal{S}=\{s_1,\dots,s_K\}$ be the
fixed vocabulary of $K=49$ sources. We define the gold source set of each query
$q$ automatically from its evidence,
\begin{equation}
S_q^\star = \mathrm{set}\big(\text{evidence sources of }q\big),
\qquad S_q^\star = \varnothing \ \text{ for null queries,}
\label{eq:gold}
\end{equation}
where the 301 null queries are mapped to the empty set because they have no
evidence. We represent each gold set as a multi-hot vector $y_q\in\{0,1\}^{K}$ whose entries are defined by $y_{qc}=\mathbf{1}[s_c\in S_q^\star]$. The evaluation metric is the set-exact
accuracy over the $N=2556$ queries,
\begin{equation}
\mathrm{Acc} = \frac{1}{N}\sum_{q=1}^{N}
  \mathbf{1}\!\left[\hat{S}_q = S_q^\star\right],
\label{eq:setexact}
\end{equation}
which is the fraction of queries whose predicted set $\hat{S}_q$ equals the gold
set exactly. We do not evaluate date filters. Although 398 baseline filters also constrain \texttt{published\_at}, the metric compares source sets only.

\subsection{Pipeline}
Figure~\ref{fig:pipeline} shows the pipeline. For a query tokenised into $T_q$
tokens, we run one forward pass through a small open-source model, loaded with
the Hugging Face Transformers library~\cite{wolf:2020} and called with
\texttt{output\_\allowbreak hidden\_\allowbreak states=True}, which gives hidden
states $h_\ell^{(t)}\in\mathbb{R}^{d}$ at layer $\ell$ for token $t$. For each layer, we store two pooled representations,
\begin{equation}
h_\ell^{\mathrm{mean}} = \frac{1}{T_q}\sum_{t=1}^{T_q} h_\ell^{(t)},
\qquad
h_\ell^{\mathrm{last}} = h_\ell^{(T_q)},
\label{eq:pool}
\end{equation}
as an $[N, n_\ell, d]$ tensor, so that the sweep over layers and poolings can run offline without recomputing the hidden states. On the pooled state $h_\ell$
of a single selected layer, a multi-label head scores each source $c$
with an independent logistic unit, and one global decision threshold $\tau$
selects the predicted sources,
\begin{equation}
p_{qc} = \sigma\!\big(w_c^\top h_\ell + b_c\big),
\qquad
\hat{S}_q = \{\, s_c : p_{qc} \ge \tau \,\},
\label{eq:head}
\end{equation}
where $\sigma$ is the logistic sigmoid and the features are standardised per
dimension. At inference, the predicted set $\hat{S}_q$ is written as a \src{}
filter, which replaces the source field produced by the prompted extractor.

\begin{figure*}[t]
\centering
\footnotesize
\resizebox{0.8\linewidth}{!}{%
\begin{tikzpicture}[
  node distance=7mm,
  box/.style={draw, rounded corners, align=center, minimum height=10mm,
              inner sep=4pt, fill=blue!4},
  llm/.style={draw, rounded corners, align=center, minimum height=10mm,
              inner sep=4pt, fill=orange!8},
  arr/.style={-{Latex[length=2mm]}, thick}
]
\node[box] (q) {Query\\\scriptsize\itshape ``\dots according to\\\scriptsize\itshape \emph{The Verge} and \emph{Wired}?''};
\node[llm, right=of q] (llm) {Small open-source LLM\\partial forward pass\\to layer $\ell$};
\node[box, right=of llm] (pool) {Mean-pool\\over tokens\\$\to h_\ell\in\mathbb{R}^{d}$};
\node[box, right=of pool] (head) {Multi-label\\head (sigmoid)\\$+$ threshold $\tau$};
\node[box, right=of head] (out) {\src{} filter\\\scriptsize\{The Verge, Wired\}};
\draw[arr] (q) -- (llm);
\draw[arr] (llm) -- (pool);
\draw[arr] (pool) -- (head);
\draw[arr] (head) -- (out);
\node[below=2mm of llm, gray, font=\scriptsize] {only first few layers};
\node[below=2mm of head, gray, font=\scriptsize] {fixed vocabulary of 49 sources};
\end{tikzpicture}%
}
\caption{The probe pipeline. A partial forward pass through the first layers of
a small open-source model produces a pooled hidden state. A multi-label
head predicts the source set from this state and writes it as a \src{} filter.
The output space consists of the 49 sources of the allow-list.}
\label{fig:pipeline}
\end{figure*}

\subsection{Layer, Pooling and Class Imbalance}
The accuracy of the probe depends on the selected layer, the pooling of token states, and the handling of class imbalance. We select the layer and the pooling by cross-validation and handle the
imbalance with standard multi-label practice, rather than setting them by hand.
With a middle layer, last-token pooling and unweighted training together, the
probe does not outperform the substring baseline.

\subsubsection{Layer}
Using iterative-stratified 5-fold cross-validation, we evaluate every layer under both poolings and pick the layer yielding the highest out-of-fold micro F1-score. Following the intermediate-layer literature~\cite{skean:2025,
lin:2025}, one would expect the best layer in the middle of the network.
However, the sweep selects one of the first layers (index 1 to 4) for all four
models (Table~\ref{tab:config}), which is consistent with source identity being
a near-lexical attribute.

\subsubsection{Pooling}
We compare mean pooling with last-token pooling on the same layers. Mean pooling
gives the higher out-of-fold F1-score for all four models and at nearly every
depth (Fig.~\ref{fig:depth}). This is consistent with the source name appearing
anywhere in the query rather than only at its end.

\subsubsection{Class Imbalance}
The empirical distribution over sources is heavily skewed. \emph{TechCrunch} appears 1190 times,
while 26 of the 49 sources appear fewer than 20 times and the rarest appears
twice. We therefore train each logistic unit by minimising a class-balanced
binary cross-entropy,
\begin{equation}
\mathcal{L}_c = -\sum_{q=1}^{N}
  \Big[\,\alpha_c^{+}\, y_{qc}\log p_{qc}
       + \alpha_c^{-}\,(1-y_{qc})\log(1-p_{qc})\Big],
\label{eq:loss}
\end{equation}
with weights inversely proportional to class frequency,
$\alpha_c^{+}=N/(2n_c)$ and $\alpha_c^{-}=N/\big(2(N-n_c)\big)$, where
$n_c=\sum_q y_{qc}$ is the support of source $c$. This weighting increases the penalty on errors for low-frequency sources. We add an L2 penalty $\frac{1}{2}\|w_c\|_2^2$ to
$\mathcal{L}_c$, with inverse strength $C=1$. We use iterative-stratified
multi-label splits~\cite{szymanski:2017} so that rare sources occur in every
fold, and report both the micro and the macro F1-score,
\begin{equation}
\mathrm{F1}_{\mathrm{micro}}
  = \frac{2\sum_c \mathrm{TP}_c}
         {2\sum_c \mathrm{TP}_c + \sum_c \mathrm{FP}_c + \sum_c \mathrm{FN}_c},
\quad
\mathrm{F1}_{\mathrm{macro}} = \frac{1}{K}\sum_{c=1}^{K}\mathrm{F1}_c,
\label{eq:f1}
\end{equation}
since the macro F1-score weights every source equally and therefore shows the
accuracy on the rare sources.

\subsection{Fixed Vocabulary and Baselines}
Restricting the output space to the 49 predefined classes prevents the probe from predicting off-list sources, whereas the generative baseline produces 128 distinct strings. We tune one global threshold $\tau$ on the out-of-fold
probabilities over a grid from 0.05 to 0.90 in steps of 0.05, maximising the
micro F1-score. The selected thresholds are 0.70 to 0.80 (Table~\ref{tab:config}). Every prediction is generated out-of-fold, ensuring that no query is evaluated by a model trained on it. The layer and the threshold are selected on the
same out-of-fold scores, so the reported numbers are slightly optimistic. As a
capacity check, we also fit an MLP with one hidden layer of 256 units on the
selected layer and pooling, with its threshold tuned in the same way.

We evaluate the probe against two baseline approaches. The first is the GPT-3.5 filter file
released with Multi-Meta-RAG. The second is a model-free substring match over
the 49 source names (case-insensitive containment). It reaches a micro F1-score of 0.957 and a macro F1-score of 0.942, and also serves as the control task for probing~\cite{hewitt:2019}.

\section{Results}
\label{sec:results}
This section presents our experimental results. We first report the selected
probe configuration of each model. Next, we benchmark the probe against GPT-3.5 and the substring baseline across all 2556 queries. Finally, we analyze where the
difference between the methods comes from, which layer is selected, and how the
accuracy depends on model size.

\begin{table*}[t]
\centering
\caption{Selected probe configuration and cross-validated F1 per model.}
\label{tab:config}
\begin{tabular}{lccccccc}
\toprule
Model & Params & Hidden states $\times$ dim & Pooling & Layer & $\tau$ & Sweep micro / macro F1 & MLP micro / macro F1 \\
\midrule
Qwen2.5-1.5B & 1.5B & $29 \times 1536$ & mean & 2\,/\,28 & 0.70 & \textbf{0.970 / 0.892} & 0.954 / 0.830 \\
SmolLM2-360M & 360M & $33 \times 960$  & mean & 4\,/\,32 & 0.75 & 0.969 / 0.885 & 0.962 / 0.840 \\
Qwen2.5-0.5B & 0.5B & $25 \times 896$  & mean & 2\,/\,24 & 0.75 & 0.967 / 0.891 & 0.962 / 0.848 \\
SmolLM2-135M & 135M & $31 \times 576$  & mean & 1\,/\,30 & 0.80 & 0.963 / 0.881 & 0.964 / 0.867 \\
\bottomrule
\end{tabular}
\end{table*}

\begin{table*}[t]
\centering
\caption{Set-exact hit accuracy by question type and overall, probe
(out-of-fold) vs.\ a model-free substring match and GPT-3.5.}
\label{tab:head2head}
\begin{tabular}{lcccccc}
\toprule
question type & Qwen2.5-1.5B & SmolLM2-360M & Qwen2.5-0.5B & SmolLM2-135M & substring & GPT-3.5 \\
\midrule
comparison           & 94.4\% & 94.5\% & 93.2\% & 92.4\% & 97.0\% & \textbf{98.2\%} \\
inference            & 91.1\% & \textbf{91.9\%} & 90.6\% & 90.9\% & 90.2\% & 91.8\% \\
temporal             & \textbf{84.2\%} & 81.8\% & 82.7\% & 82.8\% & 82.7\% & 81.8\% \\
null query           & \textbf{93.7\%} & 92.0\% & 91.4\% & 89.0\% & 66.8\% & 0.0\% \\
\midrule
\textbf{ALL (with null)}    & \textbf{90.9\%} & 90.5\% & 89.7\% & 89.4\% & 88.0\% & 80.9\% \\
\textbf{ALL (without null)} & 90.6\% & 90.3\% & 89.5\% & 89.4\% & 90.8\% & \textbf{91.7\%} \\
\bottomrule
\end{tabular}
\end{table*}

\begin{figure}[t]
\centering
\includegraphics[width=0.8\linewidth]{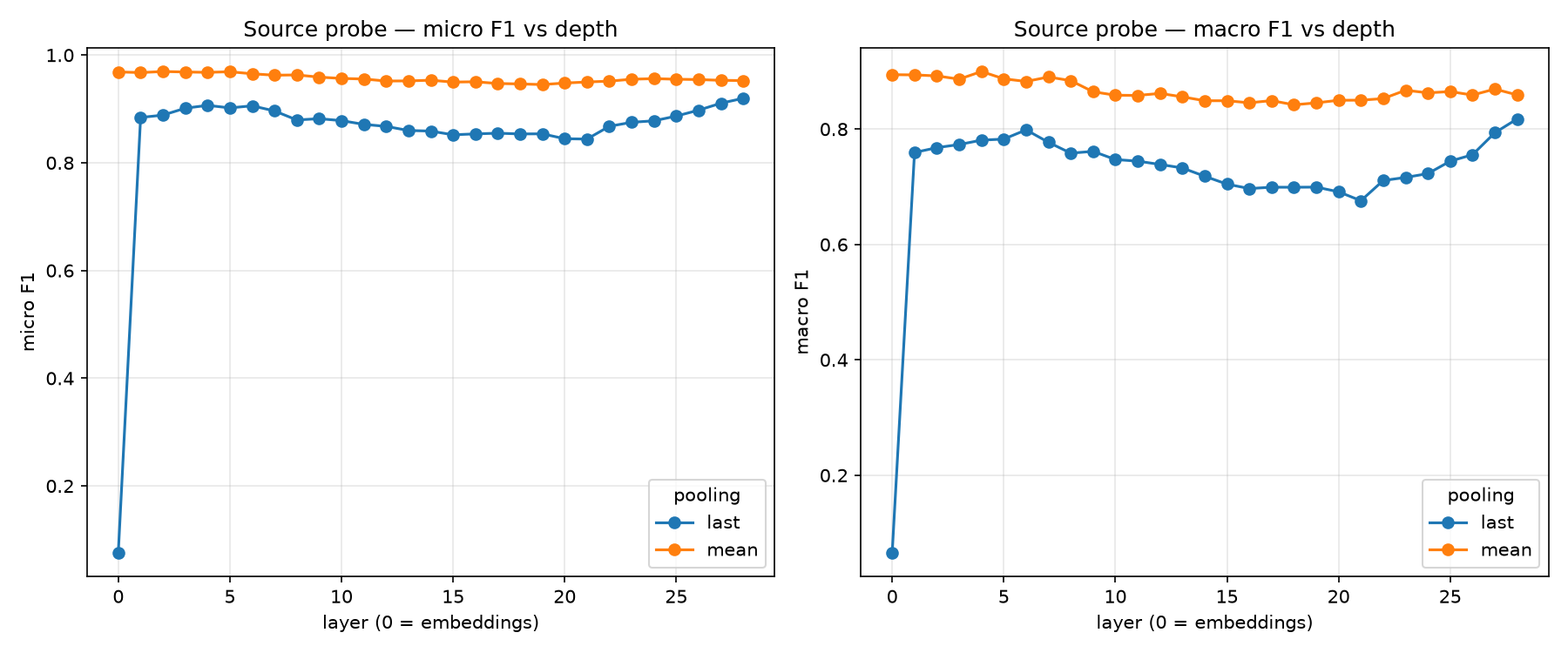}
\caption{Probe micro (left) and macro (right) F1 versus layer for Qwen2.5-1.5B,
under both poolings. Mean pooling (orange) outperforms last-token pooling (blue)
at almost every layer. Last-token pooling reaches a local peak at a shallow
layer, decreases through the middle layers and increases again only at the final
layer. Mean pooling stays high and nearly constant across layers. Neither
pooling peaks at the middle layers. Layer $0$ is the embedding layer.}
\label{fig:depth}
\end{figure}

\subsection{Probe Configuration}
Table~\ref{tab:config} reports the selected configuration of each model, chosen
by the out-of-fold micro F1-score over the sweep of class-weighted logistic
regression. The Layer column gives the selected hidden-state index ($0=$ embeddings) and the total number of transformer layers. For every model, the sweep
selects mean pooling and one of the first layers, with a hidden-state index of 1
to 4 out of 24 to 32 transformer layers. The logistic regression reaches a micro
F1-score of 0.963 to 0.970, which exceeds that of the substring baseline
(0.957). The MLP provides no meaningful improvement over logistic regression, showing that classifier capacity does not constrain the probe. The macro F1-score (0.881 to 0.892) is lower than the micro
F1-score, which shows that the remaining errors are concentrated in the rarest
sources.

\subsection{Comparison with GPT-3.5 and Substring Match}
Table~\ref{tab:head2head} compares the probe with GPT-3.5 and the substring
baseline on all 2556 queries. Across all queries, the probe achieves 90.9\% accuracy, compared to 88.0\% for the substring baseline and 80.9\% for GPT-3.5. Without
the null queries, the three methods are within 1.1 percentage points of each
other, with 90.6\% for the probe, 90.8\% for the substring baseline and 91.7\%
for GPT-3.5.

The overall difference therefore comes from the null queries. The probe correctly predicts an empty set on 93.7\% of null queries. GPT-3.5 never predicts an empty
source set and scores 0 of 301. The substring baseline returns a source for the
third of null queries that name one, and scores 66.8\%. Predicting the empty set for null queries is thus the only advantage of the probe. On the other queries,
it is on par with both baselines but does not outperform them.

\subsection{Analysis}
The results per question type (Table~\ref{tab:head2head}) show where the methods
differ on non-null queries. GPT-3.5 achieves higher accuracy on comparison queries (98.2\% compared to 94.4\%), whereas the probe achieves higher accuracy on temporal queries (84.2\% compared to 81.8\%). Overall, the probe and the two baselines extract
the sources of non-null queries with similar accuracy.

For every model, the selected layer is one of the first, with an index of 1 to 4
of the 24 to 32 transformer layers (Table~\ref{tab:config},
Fig.~\ref{fig:depth}). In contrast, prior studies identify intermediate layers as the strongest representations for multi-hop retrieval~\cite{lin:2025} and broader downstream applications~\cite{skean:2025}. A likely reason for the difference is that the source
name appears verbatim in the query in 95.4\% of gold query--source pairs. Source
identity is thus a near-lexical attribute that a partial forward pass already
exposes, whereas the tasks of prior work depend on deeper semantics.

Model size has a small effect on this task. SmolLM2-360M reaches 90.5\%,
compared to 90.9\% for the four times larger Qwen2.5-1.5B. SmolLM2-135M reaches
89.4\%, which is 1.5 percentage points less accurate than Qwen2.5-1.5B
(Table~\ref{tab:head2head}).

In practice, the probe runs locally as a partial forward pass through only 3 to 13\% of the model's layers, followed by a single linear classifier. The selected depths are 2 of 28, 4 of 32, 2 of 24 and 1 of 30
layers. This cost follows from the architecture, as the head uses a layer that
the partial pass has already computed. We did not measure latency, because our experiments cached the hidden states from full forward passes. In contrast,
the baseline requires a proprietary API call for every query and produces sources
outside the allow-list.

\section{Discussion}
In this section, we compare our results with prior work.
Table~\ref{tab:compare} summarises the comparison.

\begin{table*}[t]
\centering
\caption{Source-filter extractors compared on deployment properties and on
set-exact hit accuracy from Table~\ref{tab:head2head} (probe: Qwen2.5-1.5B).}
\label{tab:compare}
\begin{tabular}{lcccccc}
\toprule
Approach & Needs API & Output vocabulary & Null abstention & Date filters & Cost per query & Set-exact all / non-null \\
\midrule
Prompted GPT-3.5~\cite{poliakov:2024} & yes & open (128 strings) & no (0.0\%)      & yes & proprietary API call      & 80.9\% / \textbf{91.7\%} \\
Substring match                       & no  & fixed (49)         & partly (66.8\%) & no  & 49 substring tests        & 88.0\% / 90.8\% \\
Probe (ours)                          & no  & fixed (49)         & yes (93.7\%)    & no  & 1--4 layers + linear head & \textbf{90.9\%} / 90.6\% \\
\bottomrule
\end{tabular}
\end{table*}

\subsection{Comparison with LLM-Based Metadata Extraction}
Our probe directly replaces the GPT-3.5 extractor of
Multi-Meta-RAG~\cite{poliakov:2024}, an example of the metadata routing step in
advanced RAG pipelines~\cite{gao:2023}. On the same queries and allow-list, the probe is 10.0
percentage points more accurate overall but 1.1 percentage points less accurate
on non-null queries (Table~\ref{tab:compare}). Its advantages therefore lie in deployment rather than in extraction accuracy. It requires no API call, it is
deterministic, and its vocabulary is fixed. Like the small query rewriter of Ma
et al.~\cite{ma:2023}, the probe replaces an LLM call with a small model.
However, rather than using reinforcement learning guided by reader feedback, the probe is trained directly on gold labels. Constrained
decoding~\cite{geng:2023,decao:2021} would prevent sources outside the
allow-list but keep the cost of autoregressive decoding. Finally, unlike the
prompted extractor, which adds date filters to 398 queries, the probe produces
source filters only.

\subsection{Fairness of the Null-Query Comparison}
The large difference on null queries (93.7\% compared to 0.0\%) depends on our convention that a null query has an empty gold set. Moreover, the GPT-3.5 prompt
was written to extract sources, not to abstain, and a prompt that allows
abstention might reduce the difference. We therefore report the comparison
without null queries alongside the overall one.

\subsection{Comparison with Lexical and Rule-Based Extraction}
Because the source is named verbatim in most gold query--source pairs, the substring baseline serves as a proxy for span extractors ranging from string matching to compact named-entity recognizers like GLiNER~\cite{zaratiana:2024}. All of them return
a source whenever one is named in the query. The substring baseline is on par
with the probe on non-null queries, but it returns a source for every null query
that names one (Table~\ref{tab:compare}). Serving as a control task for probing~\cite{hewitt:2019}, it shows that the probe's advantage over surface matching is producing empty predictions for null queries that mention a source.

\subsection{Comparison with Layer-Selection Findings}
The selection of one of the first layers (index 1 to 4 of 24 to 32,
Table~\ref{tab:config}) differs from the middle-layer results for embedding
tasks~\cite{skean:2025} and multi-hop retrieval~\cite{lin:2025}, but it does not
contradict them. Those tasks depend on semantics that build up with depth,
whereas source identity is near-lexical. Furthermore, layer-wise probing shows that lower-level linguistic information is localized earlier in the network than semantic properties~\cite{tenney:2019}. The best layer therefore depends on the
attribute that the probe predicts, and it should be selected for each task by a
sweep.

\subsection{Comparison with Probing and Distillation Work}
Unlike probes used for interpretation~\cite{alain:2016,conneau:2018,belinkov:2022}
or monitoring~\cite{azaria:2023}, our probe is a step of the pipeline, as its
output is written directly as the database filter. A distilled student can even outperform its teacher on extraction~\cite{zhou:2024}. However, a student
trained on the GPT-3.5 filters would learn their errors, including the sources
outside the allow-list and the missing empty predictions on null queries.
Training on the gold evidence sources avoids both.

\subsection{Novelty}
Compared with the work discussed above, this paper makes four contributions. Firstly, we
present a hidden-state probe that replaces LLM source extraction in a filtered
RAG pipeline, and evaluate it with set-exact accuracy on all 2556 MultiHop-RAG
queries against both the original extractor and a lexical control. Secondly, we
find that the advantage of the probe is the empty prediction on null queries
rather than more accurate source extraction, which a single overall score would
hide. Thirdly, we show that early layers (indices 1 to 4) perform best for near-lexical extraction, in contrast to the middle-layer results for retrieval tasks~\cite{lin:2025,skean:2025}. Finally, the probe is a deterministic
extractor with a fixed vocabulary, which requires no API and uses only the first
layers of a model with 135M to 1.5B parameters.

\section{Conclusions}
We showed that a probe on the mean-pooled hidden states of one of the first
layers of a small open-source model, trained with a class-balanced loss, can
replace the source filters of the prompted GPT-3.5 extractor in
Multi-Meta-RAG~\cite{poliakov:2024}. Overall, the probe improves set-exact accuracy by 10 percentage points. This difference comes from the null queries, and on the other
queries the probe is on par with a substring baseline but does not outperform
it. Moreover, the probe requires no API call and cannot produce a source outside
the allow-list. For systems that use a probe in the pipeline, our results suggest using the earliest layer that suffices for the task. For
near-lexical attributes, this is one of the first layers, in contrast to the
middle layers reported for semantic tasks.

\subsection{Limitations}
Our study has several limitations. Firstly, our evaluation is limited to a single dataset from the news domain. Secondly, the source is a near-lexical attribute, so a
string-match baseline is already strong. Thirdly, the macro F1-score is lower
than the micro F1-score, which shows that the probe is less accurate on the
rarest sources. Finally, the experimental evaluation is based on a single cross-validation run. The splits use
one seed, so we do not report the variation between splits. Moreover, the layer
and the threshold are selected on the same out-of-fold scores that we report, so
the results are slightly optimistic.

\subsection{Future Work}
Future work includes predicting \texttt{published\_at} date filters, which are extracted by the baseline but omitted by the current probe. In addition, a hybrid of the
substring match and the probe with a gate for null queries could raise the
accuracy on non-null queries above that of string matching. Focal
loss~\cite{lin:2017} or per-class thresholds could improve the macro F1-score
on the rare sources. Finally, we plan to evaluate the impact of the probe on end-to-end retrieval metrics (MRR@10, Hits@$k$), extending evaluation beyond extraction accuracy alone.

\bibliographystyle{IEEEtran}
\bibliography{refs}

\end{document}